\documentclass[10pt,twocolumn,letterpaper]{article}

\usepackage{wacv}
\usepackage{times}
\usepackage{epsfig}
\usepackage{graphicx}
\usepackage{amsmath}
\usepackage{amssymb}
\usepackage{subcaption}

 \wacvfinalcopy 

\ifwacvfinal
 \def\assignedStartPage{1} 
\fi

\ifwacvfinal
\usepackage[breaklinks=true,bookmarks=false]{hyperref}
\else
\usepackage[pagebackref=true,breaklinks=true,colorlinks=,bookmarks=false]{hyperref}
\fi

\ifwacvfinal
\else
\fi

\begin{document}

\title{PeR-ViS: Person Retrieval in Video Surveillance using Semantic Description}

\author{Parshwa Shah\\
School of Engineering and Applied Science, Ahmedabad University\\
{\tt\small parshwa.s.btechi16@ahduni.edu.in}\\
Arpit Garg, Vandit Gajjar\\

School of Computer Science, The University of Adelaide\\
{\tt\small \{arpit.garg, vanditjyotindra.gajjar\}@student.adelaide.edu.au}

}

\maketitle

\begin{abstract}
  A person is usually characterized by descriptors like age, gender, height, cloth type, pattern, color, etc. Such descriptors are known as attributes and/or soft-biometrics. They link the semantic gap between a person’s description and retrieval in video surveillance. Retrieving a specific person with the query of semantic description has an important application in video surveillance. Using computer vision to fully automate the person retrieval task has been gathering interest within the research community. However, the Current, trend mainly focuses on retrieving persons with image-based queries, which have major limitations for practical usage. Instead of using an image query, in this paper, we study the problem of person retrieval in video surveillance with a semantic description. To solve this problem, we develop a deep learning-based cascade filtering approach (PeR-ViS), which uses Mask R-CNN ~\cite{he2017mask} (person detection and instance segmentation) and DenseNet-161 ~\cite{huang2017densely} (soft-biometric classification). On the standard person retrieval dataset of SoftBioSearch ~\cite{denman2015searching}, we achieve 0.566 Average IoU and 0.792 \%w $IoU > 0.4$, surpassing the current state-of-the-art by a large margin. We hope our simple, reproducible, and effective approach will help ease future research in the domain of person retrieval in video surveillance. The source code and pretrained weights available at \url{https://parshwa1999.github.io/PeR-ViS/}.

\end{abstract}

\section{Introduction}

\begin{figure}[h]
    \centering
    \includegraphics[width=0.50\textwidth]{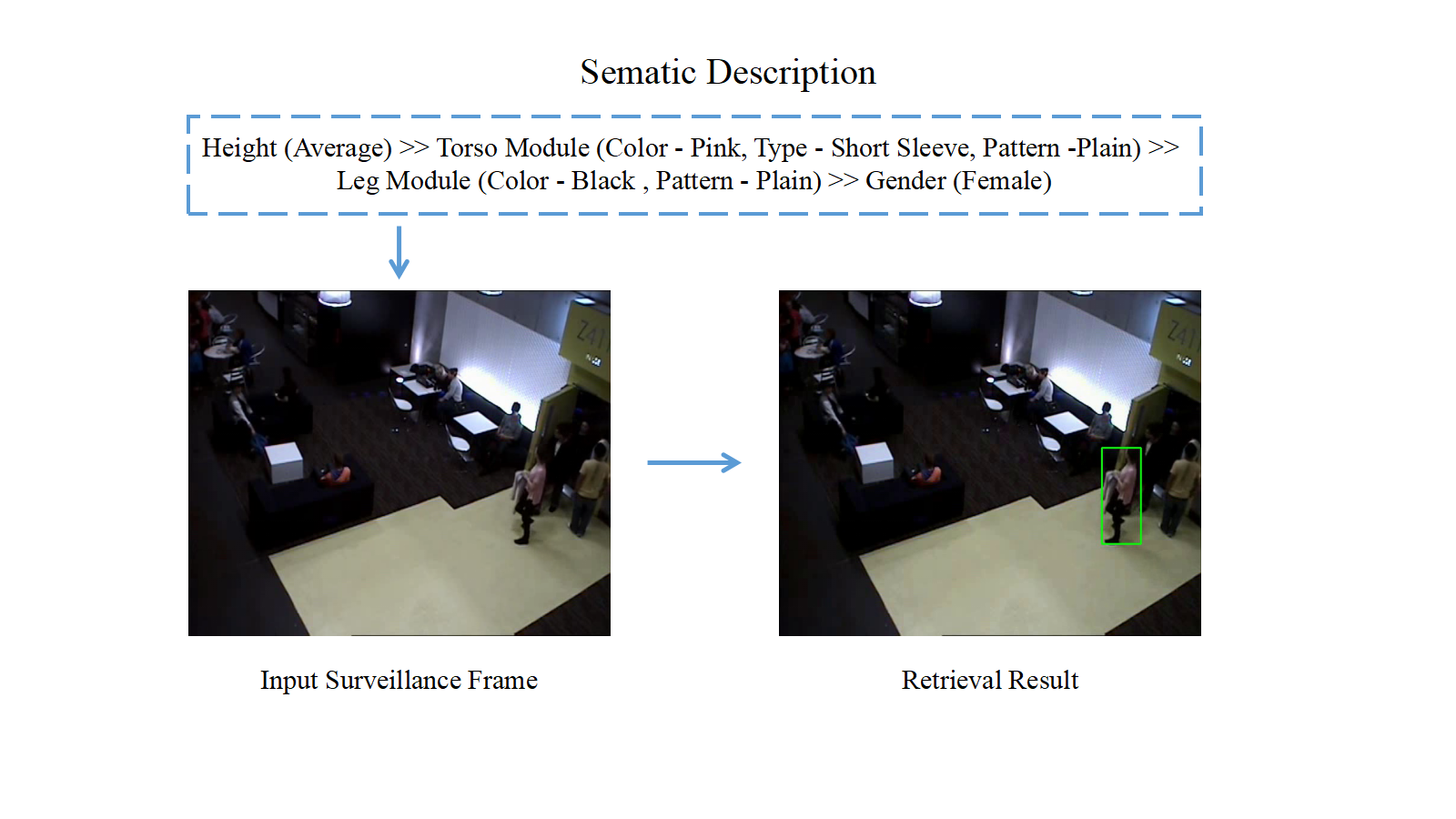}
    \caption{\textbf{Person retrieval} using a semantic description. Given the semantic description of a person, our approach PeR-ViS uses each description in a filtering mode to retrieve the correct person.}
    \label{fig:1}
\end{figure}

Recently, pedestrian attribute recognition such as age, gender, height, cloth color, and type, etc. has obtained increasing attention due to its promising outcomes in applications such as person re-identification, attribute-based person search, and person retrieval in video surveillance. Nowadays metropolitan cities are equipped with thousands of surveillance cameras, which stores a gigantic amount of surveillance data every second. To retrieve a specific person manually from large-scale videos possibly takes months to complete. Using computer vision techniques to fully automate the above task has been gathering interest within the research community. The current trend mainly solves this task based on image queries, which have major limitations and might not be suitable for practical usage. Therefore, we studied the problem of person retrieval with semantic descriptions to face these limitations. Figure \ref{fig:1} illustrates the example of the person retrieval using a semantic description.


Person retrieval with image-based queries is known as
Person Search and/or Person Re-identification in computer
vision ~\cite{zheng2011person, liao2015person, xiao2016end}. Given image query to network, it finds
the similarity between the query and that surveillance
footage. The most identical person than retrieved from the
footage according to similarity score. However, this
problem requires at least one image as a query for the
network, which has a major limitation in practice. In many
cases, such as a lost person, there might be only a
description provided of the person(s) appearance
available. Facing the limitation of image-based person
retrieval, we propose to use a semantic description for
person retrieval. It does not require a person(s) image to
be applied to the network. The semantic description can
also accurately describe the information of the person(s)
appearance.

\begin{figure*}[h!]
\begin{center}
    \includegraphics[width=1\textwidth]{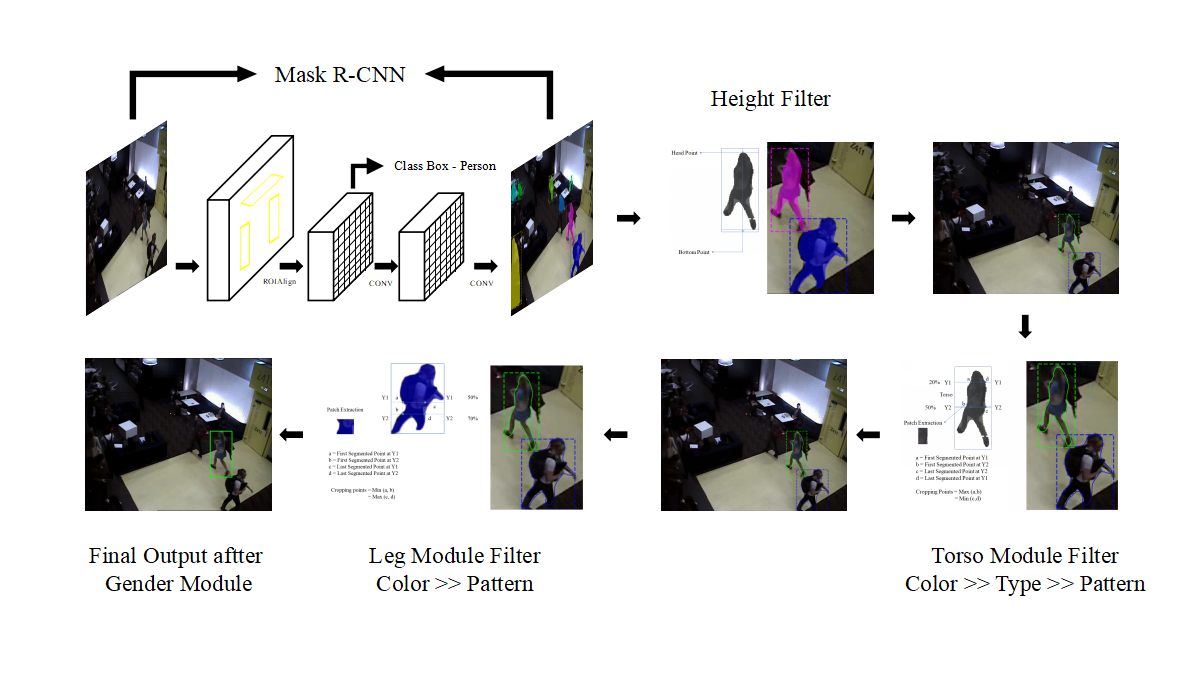}
    \caption{\textbf{Overview of the proposed approach - PeR-ViS.} We first apply Mask R-CNN to the input image for a person(s) accurate detection and instance segmentation. The detected person(s) then fed into the linear filter, which ultimately narrows down the search space and leaves only the target at the end. The filtering sequence follows the given order: Height; Torso cloth color, type, pattern; Leg cloth color, pattern; and Gender. (Best viewed in color and magnification.)}
    \label{fig:2}
\end{center}

\end{figure*}

To use this semantic description for person retrieval, we
propose PeR-ViS - a deep learning-based cascade filtering approach for person retrieval in video
surveillance. The PeR-ViS takes a semantic
description and a surveillance frame as an input and
outputs the correct retrieved person with a bounding box.
It uses Mask R-CNN ~\cite{he2017mask} for precise detection and instance
segmentation of every person in the surveillance frame. It
uses Height, Torso Module (Color $>>$ Type $>>$ Pattern),
Leg Module (Color $>>$ Pattern) and Gender as cascade
filter. These descriptors are chosen due to their
distinguishable capability. For example, the height
descriptor is view and distance invariant, while predicting
color is also invariant to angle and direction ~\cite{shah2017description}. The
height filter is designed using camera calibration
parameters, while all other filters are based on
convolutional neural networks: DenseNet-161 ~\cite{huang2017densely}. The
complete approach is illustrated in Figure \ref{fig:2}, and we will
talk about more in Section \ref{sec3}.

In summary, the main contributions of this paper are
four-fold.

\begin{itemize}
  \item We study the problem of person retrieval with the
        semantic description in video surveillance, which
        often arises in real-life scenarios, but remains wide
        open in the research community.
  \item Mask R-CNN is used by us because of following advantages
        \begin{itemize}
            \item Unnecessary background clutter is removed.
            \item As accurate segmented boundary provides precise head and feet points. Better estimation of real-world height can be made.
            \item It helps to extract accurate patch for Torso and Leg module attribute classification.
        \end{itemize}
        \item The estimated height can also be used to distinguish between the standing and sitting position of the person. This ability narrows down search space for the person of interest in the standing position.
        \item We propose a new person retrieval approach (PeR-ViS) which uses cascade filters of person(s) descriptors to narrow down the search space of detected people to leave only the target. The approach surpasses the current state-of-the-art on the SoftBioSearch dataset, achieving 0.566 Average IoU and 0.792 \%w IoU $>$ 0.4.
\end{itemize}

The rest of this paper is as follows. Section \ref{sec2} describes work-related to person search and retrieval in video surveillance. Our approach to PeR-ViS and its modules is briefly mentioned in Section \ref{sec3}. The experiment, the implementation details, and the results are described and shown in Section \ref{sec4}. Section \ref{sec5} discusses the experimental results and ablation studies. Finally, Section \ref{sec6} focuses on the possible future work and concludes the paper.


\section{Related work}
\label{sec2}
\textbf{Person Search:} A person search is a recently introduced
problem, where an image query is applied to the network
and the same/similar person can be found. Li et al. \cite{xiao2017joint}
have proposed a person search task that aims to find a
similar person(s) in the photo-gallery without bounding
box annotation. The respective data is similar to that in the
person re-identification. The major difference is that the
bounding-box is unavailable in the person search task.
Moreover, it can also be seen as a task to combine person
detection and person re-identification. There are some
other works, which try to search a person with other
modalities of data, such as attribute-based \cite{su2016deep, feris2014attribute}, and natural language-based \cite{li2017person}, which are more similar to the problem we aim to tackle in this task.

\textbf{Person re-identification:} 
Person search is indeed an extension of the Person re-identification task \cite{zajdel2005keeping, gheissari2006person}, which objective is to match a person(s) image from various cameras within a short span. The problem has drawn much attention in the computer vision research community since the last decade. Several datasets have been \cite{gou2017dukemtmc4reid, hirzer2011person, karanamsystematic, li2014deepreid, wang2016person}  proposed to tackle the task of person re-identification. Early person re-identification methods focus on manually designing distinguishable features \cite{hamdoun2008person, wang2007shape}, and learning distance metrics \cite{li2014deepreid, liao2015efficient, paisitkriangkrai2015learning}. With the growth of deep learning in recent years, many researchers have proposed several deep learning-based solutions to solve this task. Li et al. \cite{li2014deepreid} and Ahmed et al. \cite{ahmed2015improved} created CNN models for person re-identification. Both the CNNs uses a pair of cropped person(s) images as input and utilizes a binary verification loss function to train the different parameters. Ding et al. \cite{ding2015deep} and Cheng et al. \cite{cheng2016person} utilized triplet instances for training CNNs to minimize the feature distance between the similar person and maximize the distance between different people. Instead of utilizing the triplet loss function, Xiao et al. \cite{xiao2016learning} proposed to learn features by categorizing identities.

\textbf{Person Retrieval:} Before the growth of deep learning in the field of computer vision, hand-crafted methods had developed algorithms to learn part or local features. In the work of Gray et al. \cite{gray2008viewpoint}, they have partitioned person(s) into horizontal stripes to extract color and texture features. Similar partition has also been used by many other works \cite{liao2015person, prosser2010person, ma2013domain}. We also have adopted the strategy for extracting torso and leg patches accurately. Some other works utilize a more refined strategy. In the work of Gheissari et al. \cite{gheissari2006person}, authors have divided person(s) into different triangular parts for feature extraction. Cheng et al. \cite{cheng2011custom} used a genteel structure to parse the person(s) into different semantic parts. Das et al. \cite{das2014consistent} have applied histograms on the head, torso, and leg portion to extract the spatial information. The state-of-the-art on most person retrieval datasets is currently maintained by deep learning algorithms \cite{zheng2016person}. There are other methods, which are more or less similar to our work for person retrieval, however, the strategy of cascade filtering that aims to narrow down the detected people to leave only the target is our major contribution.

\section{Proposed approach - PeR-ViS}
\label{sec3}
This section introduces a deep learning-based cascade filtering approach for person retrieval in video surveillance (PeR-ViS). Figure \ref{fig:2} illustrates the complete flow diagram of PeR-ViS. Each video surveillance frame is given to state-of-the-art Mask R-CNN \cite{he2017mask} in order to detect and instantiate segment(s) of person(s). For person(s) head and feet, points are extracted for all detected and segmented person(s). For height estimation, using the camera calibration technique, it is calculated based on the height given in the semantic description. In this complete approach, height acts as a primary filter to narrow down the search space of person(s) in the frame. In the case of multiple matches, where more than one person matches the height description, additional filtering is performed using torso color, type, pattern; leg color, pattern, and gender. Instance segmentation helps us in obtaining background free extraction of the patch from the torso and leg. The number of identities is further narrow down by comparing the semantic description with the extracted patches. The preciseness of the final output is further improved by exploiting gender classification. The following sections describe the process of filter modules for person retrieval.

\textbf{Height estimation:} Person height is view-invariant, which helps to distinguish between standing and sitting position of the person. Tsai camera calibration approach \cite{tsai1987versatile} is used to estimate detected person(s) height by matching bounding box coordinates to real-world coordinates. The dataset of SoftBioSearch \cite{denman2015searching} provided 6 calibrated cameras for the calculation of real-world coordinates. Detected person(s) head and feet points are computed from the instance segmentation, which can be seen in the height filter (Figure \ref{fig:1}). Steps for Computation of height estimation are as follows:
\begin{itemize}
    \item From the given camera calibration parameters Intrinsic parameters matrix (\textbf{$I_m$}), rotation matrix (\textbf{$R_m$}) and a translation vector (\textbf{$t_v$}) are computed.
    \item The respective transformation matrix is computed as $T = I_m[R_m|t_v]$.
    \item By using radial distortion parameters, head and feet points are undistorted.
    \item Using inverse transformation of $T$ global coordinate for feet is set to $F = 0$ and global coordinates $X$, $Y$ are derived.
    \item By using $X, Y$ coordinate to compute the F coordinate of the head which also characterizes height.
\end{itemize}

Calculated height assists in narrowing down the search space within the test surveillance frame based on the semantic description (e.g. Average height (150-170 cm)). After that test surveillance frame now only contains the person(s) which matches the height’s semantic description.
During training time, annotated head and feet points are used to calculate the height from all the input surveillance frames of the video sequence. The average height ($H_{avg}$) is computed over all the surveillance frames in a given video sequence. Over the similar training video sequence, we noticed that the average height estimated from automated head and feet point is greater than was $H_{avg}$. This dissimilarity holds the wrong computation, therefore; it is subtracted from $H_{avg}$ during the testing time to equalize the error.

\begin{figure}[h]
    \centering
    \includegraphics[width=0.50\textwidth]{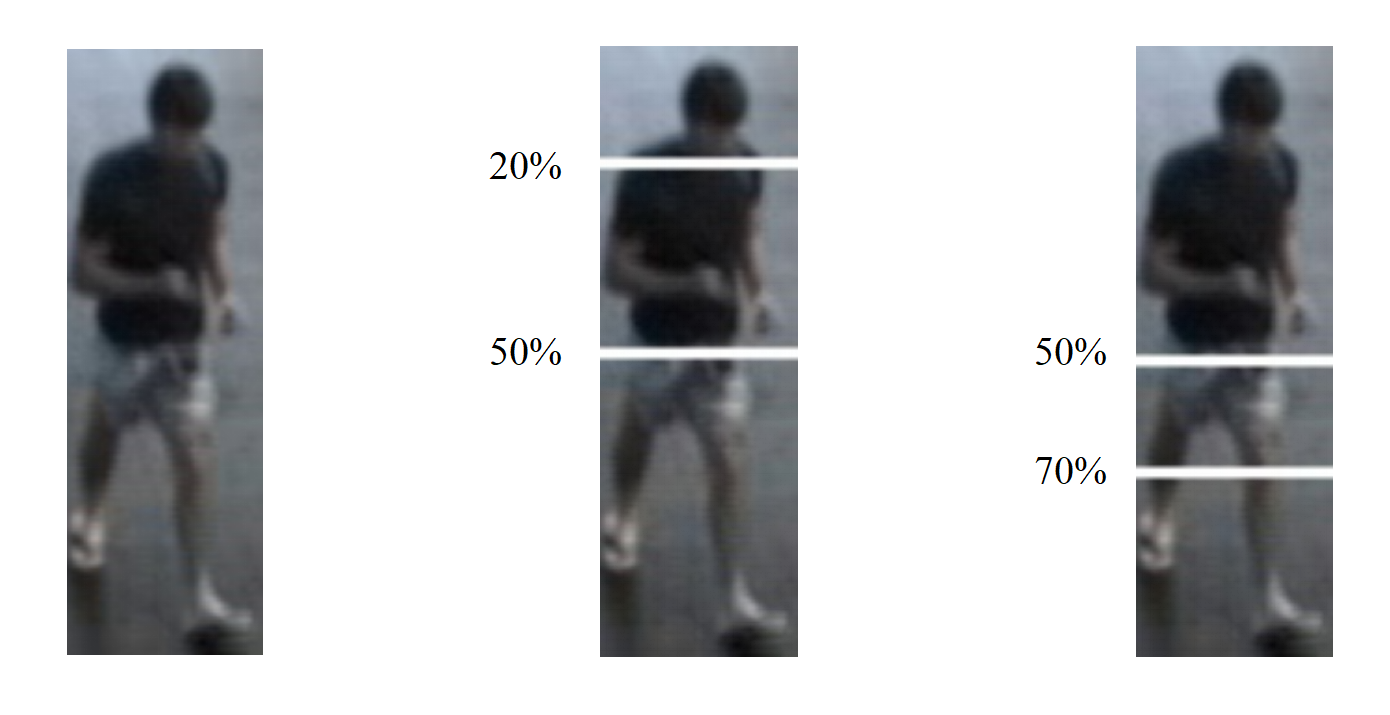}
    \caption{\textbf{Extraction of torso and leg region} from detected person(s) body.}
    \label{fig:3}
\end{figure}

\textbf{Torso and Leg Module Filter:} Mask R-CNN generates a person(s) label with a confidence score and instance segmentation. Using the golden ratio for height, the torso and leg regions are extracted. From the detected bounding box and instance segmentation, the upper 20-50\% region is classified as the torso, while the 50-70\% represents the leg portion of the person(s). The extraction of these regions is shown in Figure \ref{fig:3}. Instance segmentation is utilized to get the torso and leg portion without the background to refine the cloth color, pattern, and type classification. Patch for prediction is extracted from a region marked by points ‘a’, ‘b’, ‘c’, and ‘d’ (ref. Figure \ref{fig:1} - Torso and Leg module filter). The extracted patches are used to fine-tune DenseNet - 161 \cite{huang2017densely}, for predicting the confidence score. The SoftBioSearch dataset \cite{denman2015searching} annotation contains 13 Torso primary and secondary color, 4 torso cloth type, 8 torso cloth pattern, 13 Leg cloth primary color and secondary color, and 8 leg cloth patterns for each identity. In the case of different similar matches, the approach will refine the results by using each given modality. This feature of cascade filtering helps to narrow down the search space. Furthermore, to get and/or verify the result of the last filter (Leg cloth pattern), we use the gender filter. We noticed that in some of the cases gender filter helps to improve the performance when the surveillance frame is crowded.

\textbf{Gender Classification:} By using height and other semantic descriptions in a cascade filtering mode, the proposed approach retrieves the person for most cases. But when multiple matches come after the last filter, the approach uses Gender as the final filter for retrieving or verifying the result. Full-body images of male and female categories were used in fine-tuning DenseNet.

\section{Experiments}
\label{sec4}
This section discusses the details about the overview of the dataset, performance metric for evaluation, and implementation details. The PeR-ViS uses Mask R-CNN, which uses the pre-trained weights of Microsoft COCO \cite{lin2014microsoft} dataset for detection and instance segmentation. The Mask R-CNN model uses the ResNet-101 FPN as the backbone architecture, which has achieved Average Precision (AP) of 35.7 on the COCO test-dev set.

\subsection{Dataset overview}
\label{sec4_1}
Our work uses the SoftBioSearch database \cite{denman2015searching}, which
uses 6 stationary calibrated cameras and consists of 110
unconstrained training video segments. Each of these 6
cameras is annotated with Tsai’s camera calibration
technique \cite{tsai1987versatile}. Every video sequence in the training set is
labeled with a set of semantic descriptions to describe the
person(s) identity. In addition, with the provided
description, nine key-points were also annotated in the
case to use human key-point estimation and exploits with
the person retrieval task. More precise details on the
camera calibration and human body key-point can be
found in \cite{denman2015searching}.

The test set consists of further 41 person identities taken
from 4 of the 6 cameras (Here camera numbers 1 and 6 are
not used) used in collecting the training set. Compilation
and annotation of the test set follow the same instructions
of the train set to ensure parity, with at least the first 30
surveillance frames of each video sequence reserves to
allow the person(s) identity to fully enter in the camera
view.

For more details on the performance evaluation, the test
person identities were separated into very easy, easy,
medium, and hard.

The given labels are defined below:
\begin{itemize}
    \item Very Easy: randomly populated scenario, no complex factors, and target person identity clearly visible.
    \item Easy: Scenario consists of one or more people, but the person(s) identity is clearly distinguishable.
    \item Medium: From the following factors one of the factors will be present: Similar type of the identity present in the scenario, Very heavy occlusion with the target, a crowded scene.
    \item Hard: Two or more of the above factors present in the scenario.
    In the test set out of 41 person identities, 6 are labeled as very easy, 13 as easy, 12 as the medium, and 10 as hard.
\end{itemize}

\subsection{Performance metric}
As described in SoftBioSearch dataset, the metrics use
the Intersection Over Union ($IoU$) given by Eq. 1. An
$IoU_{avg}$ is calculated per video sequence, and video
sequences results are averaged over all video sequences to
obtain a final accuracy measure.
$IoU = \frac{D \cap GT}{D \cup \ GT}$
Where D is the obtained bounding-box from the approach and GT is the Ground Truth bounding box.

\subsection{Implementation details}
The fine-tuning is accomplished on a workstation with
an Intel Xeon core processor and accelerated by NVIDIA
TitanX 12 GB GPU. All experiments run in Tensorflow
1.8 \cite{abadi2016tensorflow}.

\textbf{Data Augmentation:} In order to achieve generalization of training data for improved performance and robustness data augmentation is used. E.g. In the
training set after removing the surveillance frames with
partial occlusion, the final set contains 8577 images from
110 subjects. Thus training DenseNet with only 8577
images may result in over-fitting, which is avoided using a
data augmentation scheme. Each training frame is
horizontally and vertically flipped, rotated with 10 angles
{$1^o$,$2^o$, $3^o$, $4^o$, $5^o$, $-1^o$, $-2^o$, $-3^o$, $-4^o$, $-5^o$} and brightness
increased with a gamma factor of 1.5.

\textbf{DenseNet training for Cloth Color, type, pattern; and
gender:} Cloth Color, Type, Pattern, and gender models are fine-tuned using DenseNet - 161 which is pre-trained on the ImageNet dataset.
The SoftBioSearch dataset consists of 1704 patches
divided into 13 torso and leg primary and secondary
colors; 8 torso and leg patterns; and 4 torso type. Extra
patches for training these attributes are extracted using 4
human key-points provided in annotations (Left-right
shoulder and left-right waist). In order to deal with light
changes, these patches are augmented by increasing the
brightness with a gamma factor of 1.5. Thereafter, almost
17000 patches are generated and further divided into
80-20\% train and validation set.

All the networks for the torso and leg module are trained using mini-batch stochastic gradient descent (SGD). Due to the computation cost, the fine-tuning strategy is the same for all the descriptors. The networks were trained for 20 epochs with a learning rate is set to 0.001, dropout set to 0.35, and effective mini-batch size is set to 32. Table \ref{tab:1} shows the validation accuracy of different descriptors.

For the gender classification, the initial data augmentation generated 105980 images for training gender descriptor, which is almost 13 times larger than the original training set (8577). 20\% of total images were used for validation. For the gender fine-tuning, the network was trained for 30 epochs with a learning rate of 0.01, dropout rate of 0.25, and effective batch size is set to 64.

\begin{table}
\begin{center}
\begin{tabular}{|l|c|}
\hline
\textbf{Descriptor} & \textbf{Validation Accuracy} \\
\hline\hline
Torso Color & 81.29\% \\
Torso Type & 79.14\% \\
Torso Pattern & 76.5\% \\
Leg Color & 71.52\% \\
Leg Pattern & 72.5\% \\
Gender & 77.79\% \\
\hline
\end{tabular}
\end{center}
\caption{\textbf{Accuracy results} on the validation set of
different semantic descriptor based on the above
hyper-parameter setting.}
\label{tab:1}
\end{table}

\section{Experimental results}
\label{sec5}
This section covers the qualitative and quantitative experimental results derived from a test set of 41 person identities. Overall results are shown in Table \ref{tab:2}, including a comparison with the baseline algorithm and current state-of-the-art methods. Form this we can see that our approach - PeR-ViS outperformed the current state-of-the-art by a large margin.

\begin{table}[h!]
\begin{center}
\begin{tabular}{|l|c|c|}
\hline
\textbf{Approach} & \textbf{Average IoU} & \textbf{\%w $IoU > 0.4$} \\
\hline\hline
Baseline \cite{denman2015searching} & 0.290 & 0.669\\
Galiyawala et al. \cite{galiyawala2018person} & 0.363 & 0.522\\
Schumann et al. \cite{schumann2018attribute}& 0.503 & 0.759\\
Yaguchi et al. \cite{yaguchi2018transfer} & 0.418 & - \\
Yaguchi et al. \cite{yaguchi2018transfer} & 0.462 & - \\
Yaguchi et al. \cite{yaguchi2018transfer} & 0.511 & 0.669\\
\textbf{Ours} & \textbf{0.566} & \textbf{0.792} \\
\hline
\end{tabular}
\end{center}
\caption{\textbf{Overall IoU} of different methods on the test set.}
\label{tab:2}
\end{table}

\begin{figure}[h!]

    \begin{subfigure}{0.5\textwidth}
    \centering
    \includegraphics[width=1\textwidth]{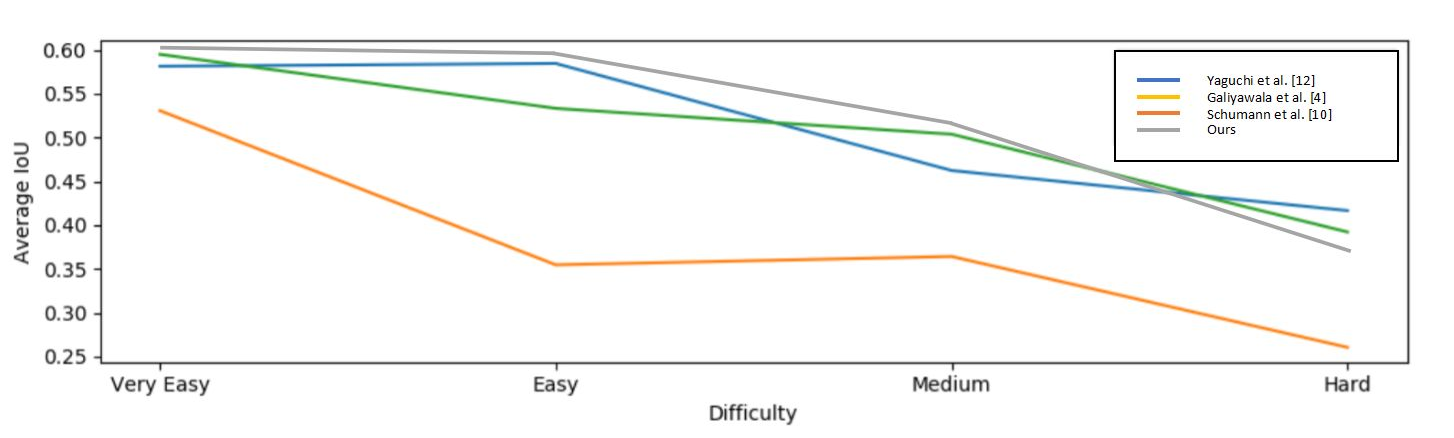}
    \label{fig:4a}
    \end{subfigure}
    
    \begin{subfigure}{0.5\textwidth}
    \centering
    \includegraphics[width=1\textwidth]{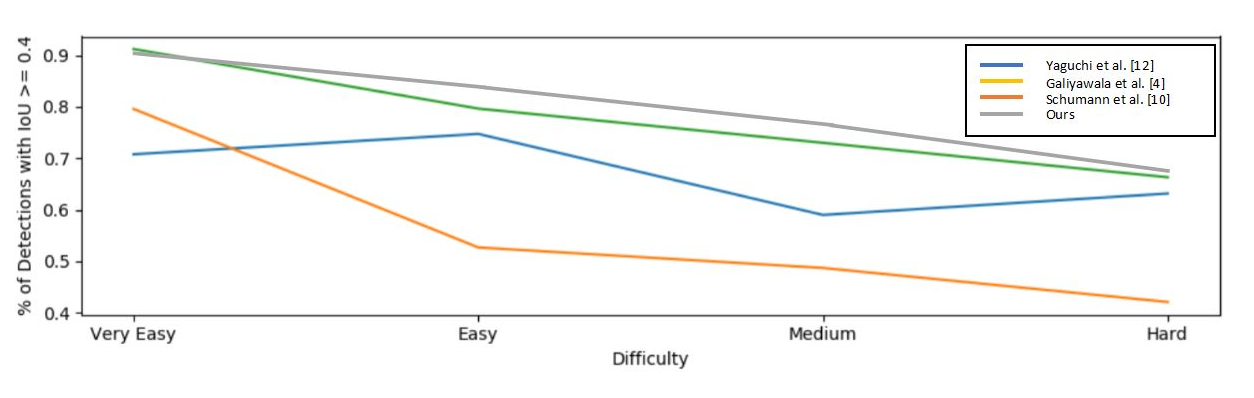}
    \label{fig:4b}
    \end{subfigure}
    \caption{\textbf{Performance broken down.} (a) by sequence difficulty and compared with Average $IoU$. (b) by sequence difficulty and compared with \%w $IoU > 0.4$. (from top to bottom - a, b)}
    \label{fig:4}
\end{figure}

Considering the algorithms used, \cite{yaguchi2018transfer}, \cite{galiyawala2018person}, and \cite{schumann2018attribute} uses deep learning method that deploys a CNNs for detecting person(s) identity in the surveillance frame. The baseline of \cite{denman2015searching} uses an avatar, which is a non-deep learning approach, that is constructed from the input semantic query to drive a gradient descent search. It is noticeable that both \cite{galiyawala2018person} and \cite{denman2015searching} uses very few descriptors, while \cite{yaguchi2018transfer} and \cite{schumann2018attribute} uses the full set of available descriptors, likely improving the performance.

\begin{figure*}[h!]
    \begin{center}
    \includegraphics[width=1\textwidth]{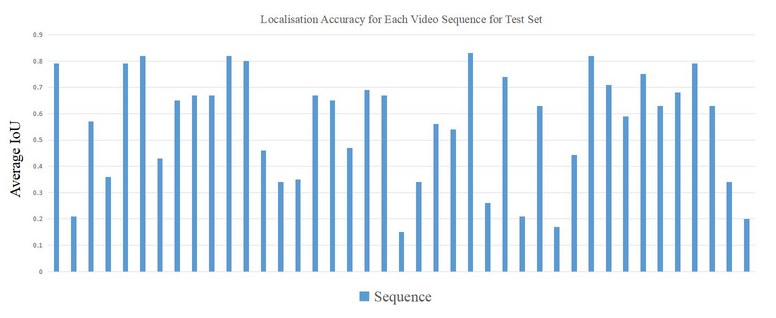}
    \end{center}
    \caption{\textbf{Per Sequence Performance} for test set.}
    \label{fig:5}
\end{figure*}

To further analyze performance, we break the results by
difficulty level (Figure \ref{fig:4}) described in Section \ref{sec4_1} and into individual video sequences (Figure \ref{fig:5}). From Figure \ref{fig:4} it is clear performance decrease with an increase in difficulty. The “Very Easy” video sequences mainly contain only a single moving person identity, and as seen in the figure all
other methods perform well on these video sequences. As the difficulty level increases, \cite{galiyawala2018person} suffers a very much decrease in the performance compared to others. A very slight difference can be spotted in the performance of \cite{yaguchi2018transfer} and \cite{schumann2018attribute}. Both achieve very similar IoU’s, but as compared with our approach their algorithms fail to perform well in the hard video sequences. Thus indeed our approach performs very well on medium and hard sequences and breaks the current state-of-the-art by a
margin of more than 5\%.
Figure \ref{fig:5} shows the performance of PeR -ViS on each
video sequence. It is noticeable that performance varies
across each video sequence and several sequences pose a
challenge for retrieval. From Figure \ref{fig:4} as compared with
\cite{schumann2018attribute}, the authors use a tracking approach, which helps to
reduce the error that may impact non-tracking approaches
such as ours and others. From this, it is noticeable that
ours and \cite{yaguchi2018transfer} have the inclination to either detect a
person’s identity very precisely or very badly.

\begin{figure*}[h!]
    \begin{subfigure}{1\textwidth}
    \centering
    \includegraphics[width=0.5\textwidth]{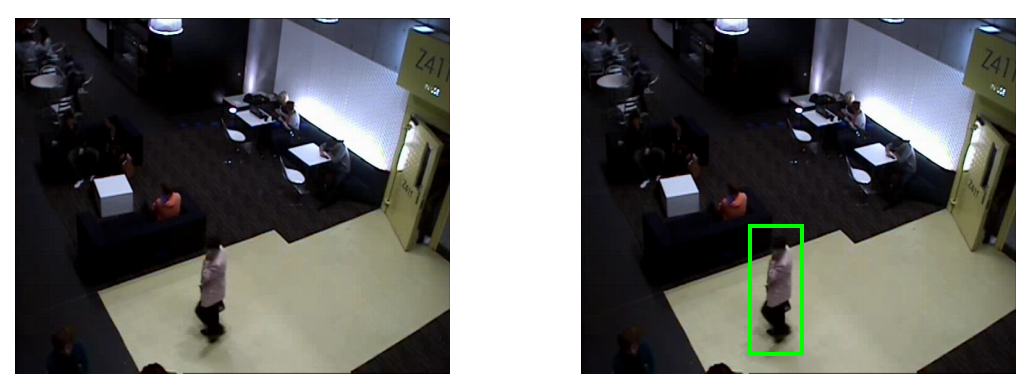}
    \caption{Person retrieved using only height.}
    \label{fig:6a}
    \end{subfigure}
    
    \begin{subfigure}{1\textwidth}
    \centering
    \includegraphics[width=0.75\textwidth]{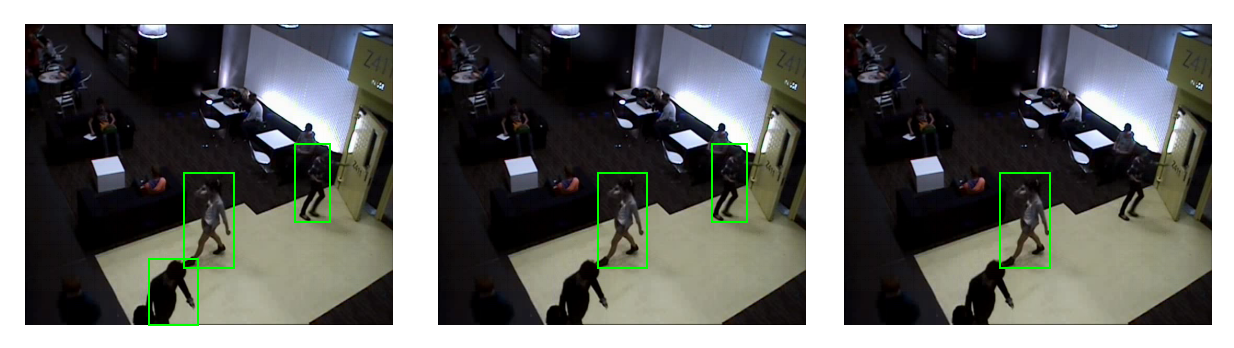}
    \caption{Person retrieved using height and torso module filter.}
    \label{fig:6b}
    \end{subfigure}
    
    \begin{subfigure}{1\textwidth}
    \centering
    \includegraphics[width=1\textwidth]{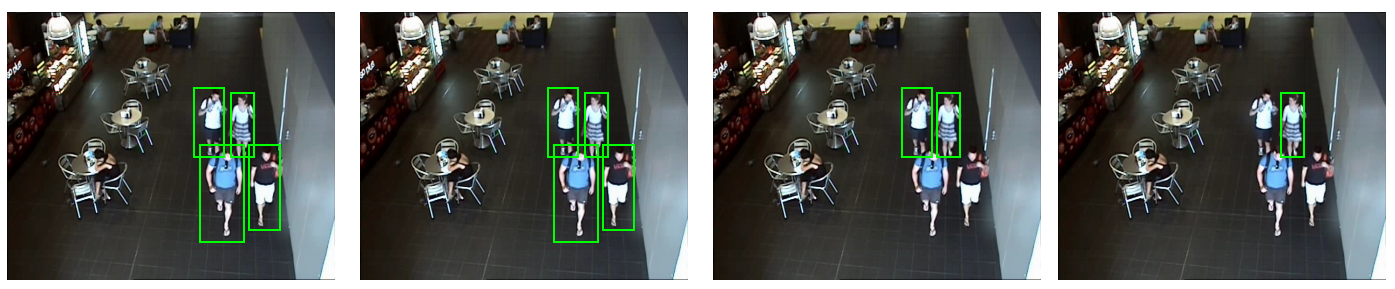}
    \caption{Person retrieved using height, torso, leg and gender module filter.}
    \label{fig:6c}
    \end{subfigure}
    
    \caption{\textbf{True positive cases} of person retrieval with semantic description. (Best viewed in color and magnification.)}
    \label{fig:6}
\end{figure*}

\begin{figure*}[h]
    \begin{center}
    \includegraphics[width=1\textwidth]{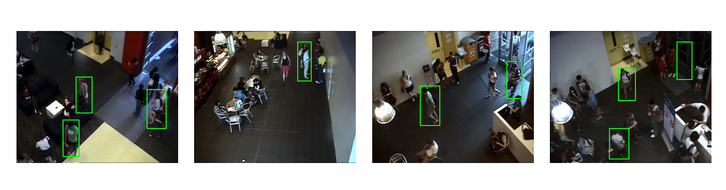}
    \end{center}
    \caption{\textbf{False negative cases of person retrieval.}(a) incorrect color classification with multiple person(s), (b) multiple person(s) with occlusion, (c) multiple person(s) with same torso color and height class and (d) person detection fails. The supplementary material contains more true positive results. (from left to right – a, b, c, d) (Best viewed in color and magnification.)}
    \label{fig:7}
\end{figure*}

Surprisingly, two of the video sequences (20 and 39)
contain very less crowd, but very complicated by having a
very similar person identity. The other two video
sequences consist high amount of crowd, and we see
almost all the systems are having a decrease in performance. To further showcase the person retrieval
results, Figure \ref{fig:6} and Figure \ref{fig:7} show some of the examples
of True positive and False negative cases based on
semantic descriptions.

Figure \ref{fig:6} shows true positive cases of person retrieval
using semantic descriptions. In Figure \ref{fig:6}, images from left
to right indicate the input test frame, the output of the
height module filter, the output of Torso and Leg module
filter and gender module filter respectively. Figure \ref{fig:7} shows
the results when the approach fails to retrieve the person correctly. It also
indicates that the dataset is created in challenging
conditions. In Figure \ref{fig:7}, the approach fails due to following
results: (a) Multiple persons with the same torso color yields incorrect color classification,
(b) Multiple person(s) with occlusion, (c) Same height
class appears when multiple persons comes in the
surveillance frame and (d) Mask R-CNN fails for person
detection.

\subsection{Ablation experiments}
We run some ablations to analyze our PeR-ViS approach \\

\textbf{Choice of classification network:} To test the influence of
a classification network on the proposed approach, we
have formed five video sequences from test set consists of
Easy, Medium, and Hard category. We have used
AlexNet, VGG-16, and ResNet-152 for ablation
experiment and average IoU is used for evaluation. All the
networks perform well on the easy video sequences. Next
on the medium sequences, where background clutter and
less occlusion are present, AlexNet and VGG-16 perform
poorly. Here ResNet-152 shows equivalent performance to DenseNet-161. The true performance of DenseNet-161 is noticed in the hard category, where it achieves
excellent performance. The Table \ref{tab:3} shows the result of
IoU using different networks.

\begin{table}[h!]
\begin{center}
\resizebox{0.5\textwidth}{!}{
\begin{tabular}{|l|c|c|c|c|c|}
\hline
 & \textbf{Very Easy} & \textbf{Easy} & \textbf{Easy} & \textbf{Medium} & \textbf{Hard} \\
\hline\hline
AlexNet & 0.567 & 0.534 & 0.523 & 0.325 & 0.183 \\
VGG-16 & 0.641 & 0.621 & 0.615 & 0.336 & 0.237 \\
ResNet-152 & 0.742 & 0.712 & 0.64 & 0.492 & 0.36\\
DenseNet-161 & 0.762 & 0.733 & 0.642 & 0.582 & 0.461\\
\hline
\end{tabular}}
\end{center}
\caption{\textbf{This table shows the IoU result}on 5 video
sequences (Sequence Number 4, 13, 21, 23, and 28) using
different network architecture in our approach.}
\label{tab:3}
\end{table}

\textbf{More descriptors:} In our work, we have used Height;
Torso and Leg module; and Gender descriptors. The
choice of these descriptors is purely based on perceptive
ability. We have tried to add more descriptors such as
Age, Shoe color, Luggage, and Human body’s build but
none of the modality was capable to improve the
performance reported in our work.

\section{Discussion and Conclusion}
\label{sec6}
The proposed approach - PeR-ViS retrieves the
person in video surveillance based on the semantic
description of Height; Torso Cloth color, type, and pattern;
Leg Cloth color and pattern; and Gender. The major
benefit from this filtering sequence is that Height, and
Torso Color, Type, and Pattern are easily differentiable
compare to other descriptors, where the heavy crowd is
present, leg patch won’t easily extract. Thus the choice of
this filtering sequence is most important in our work.
Also, instance segmentation allows precise height
estimation and accurate color patch extraction from the
torso and leg. Thus, our algorithm achieves an average
IoU of 0.566 and \%w $IoU > 0.4$ of 0.792, surpassing the
current state-of-the-art by a large margin. We have also
provided the code snippet up-to Torso type filter (Height
$>>$ Torso Cloth Color $>>$ Torso Cloth Type and Gender
for verification). The possible future work will now focus
on how to improve the results by incorporating human
pose estimation and other descriptors and investigate the
architecture for generalization in real-world scenarios.
Furthermore, the tracking approach is also useful when the
person is retrieved with Height and/or Torso module filter,
which will essentially lower the computation usage.

\section*{Acknowledgements}
{We would like to thank anonymous reviewers for providing us their valuable feedback on our paper. We would like to express our deep gratitude to Dr. Mehul S. Raval, Dr. Hiren Galiyawala and Mr. Kenil Shah for providing useful comments and discussion. We would also like to thank Ms. Ayesha Gurani, Mr. Viraj Mavani and Mr. Yash Khandhediya for their help with manuscript}




{\small
\bibliographystyle{ieee_fullname}
\bibliography{egbib}
}

\end{document}